\documentclass{article}

\usepackage{microtype}
\usepackage{graphicx}
\usepackage{booktabs}
\usepackage{hyperref}
\usepackage[table]{xcolor}

\usepackage[accepted]{icml2026}

\usepackage{amsmath}
\usepackage{amssymb}
\usepackage{mathtools}
\usepackage{amsthm}
\usepackage[capitalize,noabbrev]{cleveref}

\usepackage{algorithm}
\usepackage{algorithmic}
\usepackage{placeins}  % For \FloatBarrier to keep tables in section

\theoremstyle{plain}

\theoremstyle{definition}

\theoremstyle{remark}

\newcommand{\ours}{\textsc{SubKG-Agent}}

\newcommand{\benchmark}{NuScale-MQA}

\icmltitlerunning{LLM-Guided Planning for Multi-hop Nuclear Regulatory Document Exploration}

\begin{document}

\twocolumn[
  \icmltitle{LLM-Guided Planning for Multi-hop Reasoning \\
    over Multimodal Nuclear Regulatory Documents}

  \icmlsetsymbol{equal}{*}

  \begin{icmlauthorlist}
    \icmlauthor{Mingyu Jeon}{equal,mod}
    \icmlauthor{Bokyeong Kim}{equal,kaeri,ust}
    \icmlauthor{Suwan Cho}{mod}
    \icmlauthor{Jae Young Suh}{mod}
    \icmlauthor{Yonggyun Yu}{kaeri,ust}
  \end{icmlauthorlist}

  \icmlaffiliation{mod}{MODULABS, Seoul, Republic of Korea}
  \icmlaffiliation{kaeri}{Korea Atomic Energy Research Institute, Daejeon, Republic of Korea}
  \icmlaffiliation{ust}{University of Science \& Technology, Daejeon, Republic of Korea}

  \icmlcorrespondingauthor{Yonggyun Yu}{ygyu@kaeri.re.kr}

  \icmlkeywords{Retrieval-Augmented Generation, Knowledge Graphs, Agentic AI,
    Nuclear Regulatory Documents, Multi-hop Reasoning, Planning, Multimodal AI}

  \vskip 0.3in
]

\printAffiliationsAndNotice{\icmlEqualContribution}

% === Sections (matching Agentic_tree_search/research_paper/) ===
%-----------------------------------------------------------------------
\begin{abstract}
Reviewing nuclear regulatory documents requires multi-hop reasoning across tens of
thousands of pages, where judgments depend on evidence assembled across multiple
chapters. We frame this task as planning: an LLM-based agent observes the
evidence collected so far, picks the next document fragment to inspect, and stops when
the evidence is sufficient. The agent operates over a vectorless document tree using
browse, read, and search tools, and maintains a dynamic knowledge graph (KG) as state.
On a 200-question benchmark over NuScale Final Safety Analysis Report (FSAR) documents,
the system reaches \textbf{81.5\%} accuracy with a RAGAS Faithfulness of \textbf{0.93}.
The dominant performance factor is planning: against PageIndex, which uses the same
document tree without state-conditioned action selection, the gap is \textbf{+38.0pp} (43.5\% to
81.5\%, $p<0.001$). The system also outperforms LightRAG (73.0\%, $p<0.05$), HippoRAG
(70.5\%, $p<0.01$), and GraphRAG (49.5\%, $p<0.001$), and matches RAPTOR (75.5\%,
$p=0.11$) without offline indexing. Edge inference adds 2.8$\times$ cost without raising
accuracy; we retain it as a traceability module. Of 7{,}391 inferred edges, 3
\textsc{Violates} edges (0.04\%) flag scope boundaries (Q058) and partial conformance
(Q176) as typed annotations that a human reviewer can audit.
\end{abstract}
          % title_abstract.md
%--- Introduction (compact for 8-page limit)
\section{Introduction}
\label{sec:intro}

LLM-based agents have advanced rapidly in general-purpose domains through paradigms such as ReAct \cite{react2023}, Toolformer \cite{toolformer2023}, and Self-RAG \cite{selfrag2024}. Safety-critical regulatory domains impose constraints these paradigms do not address. Osprey \cite{osprey2026} identifies the absence of action pre-visibility in existing frameworks. Lee \cite{lee2025} analyzes the conflict between LLM opacity and the quality assurance traceability requirements of 10 CFR (Code of Federal Regulations) 50 Appendix B. Nuclear regulatory document review, which examines Final Safety Analysis Reports (FSARs) and renders conformance judgments to enable plant licensing, is where these constraints bind most tightly.

FSAR review has three properties that distinguish it from single-shot question answering. First, judgments depend on evidence accumulated from multiple chapters: answering ``Does the Emergency Core Cooling System (ECCS) design satisfy 10 CFR 50.46(b)?'' requires specifications from Chapter 5 cross-referenced against requirements in Chapter 1, with the verdict synthesized from both. Second, evidence is multimodal: specification tables and engineering drawings co-determine answers alongside the prose. Third, the review process includes a sufficiency judgment in which the reviewer decides when collected evidence is enough to render a verdict. Existing RAG (Retrieval-Augmented Generation) methods address none of these. Chunking severs cross-references \cite{ragsurvey2023}. Graph-based methods (GraphRAG \cite{graphrag2024}, LightRAG \cite{lightrag2024}) build static global graphs and lack an iterative judgment loop. All require offline indexing that is incompatible with continuously revised FSARs.

Our key observation is that regulatory review is a \textbf{planning problem}: given a goal (regulatory judgment), state (evidence collected), and actions (document navigation), the reviewer must decide what to examine next. This is a sequential decision-making process irreducible to single-shot retrieval. We construct the document as a \textbf{text-based environment} with browse/read/search actions, enabling a closed planning loop, where 33\% of queries terminate early and 67\% use the full 4-hop budget.

We make four contributions. (1) \textbf{Document as environment for planning}: a state-conditioned planning loop over a vectorless document tree, isolated against PageIndex (same environment, no planning) by a $+38.0$pp accuracy gap ($p<0.001$). (2) \textbf{Multimodal evidence handling integrated into the planning loop}: vision processing applied at the answer step yields $+18$pp over RAPTOR on table-only questions while keeping all intermediate operations text-only. (3) A \textbf{200-question nuclear regulatory benchmark} along three orthogonal axes (reasoning type, evidence complexity, modality) that jointly determine the regulatory review process. (4) \textbf{Traceability analysis via \textsc{Violates} case study}: post-retrieval edge inference adds 2.8$\times$ cost with no accuracy gain, but produces auditable reasoning paths. Among 7{,}391 inferred edges, only 3 (0.04\%) are \textsc{Violates}, identifying scope exclusions (Q058) and partial conformance (Q176) as typed annotations that natural-language answers cannot represent structurally. This satisfies 10 CFR 50 Appendix B traceability requirements.
      % introduction.md  (Section 1)
%--- Related Work (compact for 8-page limit)
\section{Related Work}\label{sec:related}

\paragraph{RAG and graph-based retrieval.}
Standard RAG \cite{rag2020} fragments context through chunking. GraphRAG \cite{graphrag2024} and LightRAG \cite{lightrag2024} construct global knowledge graphs but require expensive pre-indexing and generic edge schemas. RAPTOR \cite{raptor2024} builds recursive abstractive trees; HippoRAG \cite{hipporag2024} uses hippocampal memory models. All perform single-shot retrieval without iterative evidence accumulation.

\paragraph{Agentic information retrieval.}
Iterative retrieval methods such as IRCoT, FLARE, and Iter-RetGen interleave retrieval with chain-of-thought generation but lack persistent state and dynamic termination. Self-RAG \cite{selfrag2024} inserts reflection tokens for adaptive retrieval within a single pass. PRISM \cite{prism2025} separates precision and recall through an iterative Selector--Adder agent loop. APEX-Searcher \cite{apexsearcher2026} combines reinforcement learning (RL) with supervised fine-tuning (SFT) for planning but requires training. For document navigation, ReadAgent \cite{readagent2024} uses gist memories, DocAgent \cite{docagent2025} extracts XML outlines, and BookRAG \cite{bookrag2025} routes queries through hierarchical indices. PageIndex \cite{pageindex2024} builds vectorless trees but operates as a single-pass tool. Our work departs by constructing the document as a \textbf{scalable text-based environment} with persistent KG state, a planning loop, and dynamic termination, in a \textbf{training-free} setting.

\paragraph{Planning, world models, and KG-RAG.}
ReAct \cite{react2023}, Tree of Thoughts \cite{tot2023}, and LATS \cite{lats2024} establish LLM-based planning in PDDL/robotic/web environments through tree search with reflection; we extend this paradigm to \textbf{information environments} over structured documents. GWM \cite{gwm2025} uses graph-structured state with message-passing; we similarly employ a Dynamic Sub-KG but generate explicit relational edges through LLM inference rather than embeddings. Our edge ontology draws on SysML traceability \cite{sysml2014}, argumentation mining \cite{argmining2013,textual_entailment2012}, causal KG \cite{causal_kg2019}, and prerequisite learning \cite{prerequisite2017}.

\paragraph{Nuclear NLP and evaluation.}
NuclearQA \cite{nuclearqa2023} and NukeBERT \cite{nukebert2020} address single-hop factual extraction. Our benchmark is the first to target multi-hop, multimodal, cross-chapter regulatory judgment. We adopt dual evaluation: RAGAS \cite{ragas2024} for grounding quality and LLM-as-Judge \cite{llmasjudge2023} with three-evaluator majority vote.
      % related_works.md (Section 2)
%--- Method section, converted from Agentic_tree_search/research_paper/method.md
\section{Method}
\label{sec:method}

\begin{figure*}[t]
  \centering
  \includegraphics[width=\textwidth]{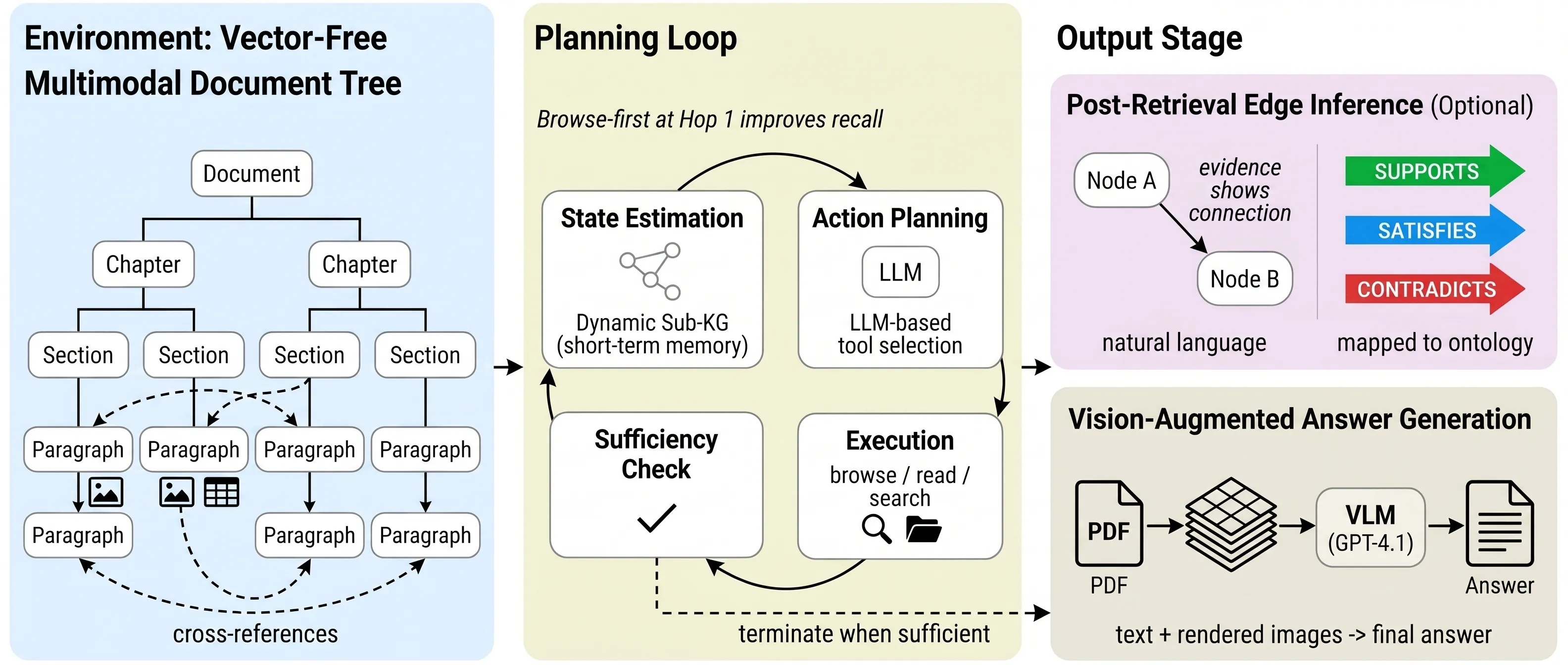}
  \caption{Overall architecture. The vectorless document tree (left) serves as the environment. The planning loop (center) iterates through state estimation, action planning, execution, and sufficiency checking. Post-retrieval edge inference and vision-augmented answer generation (right) are applied at the output stage.}
  \label{fig:pipeline}
\end{figure*}

\subsection{Problem Formulation: Regulatory Review as Planning}
\label{sec:formulation}

We formulate regulatory document exploration as a planning problem $\langle \mathcal{S}, \mathcal{A}, f_\mathrm{tr}, \phi \rangle$ with a single agent acting over a structured information environment:

\begin{itemize}
  \item \textbf{State} $s_t \in \mathcal{S}$: the Dynamic Sub-Knowledge Graph $\mathcal{G}_t = (\mathcal{V}_t, \mathcal{E}_t)$ collected through hop $t$, representing the agent's current evidence and inferred relationships (\S\ref{sec:state}).
  \item \textbf{Action} $a_t \in \mathcal{A}$: a tool invocation from $\{\texttt{browse}(d, v),\, \texttt{read}(d, v),\, \texttt{search}(\kappa)\}$ over documents $d$, tree nodes $v$, and keywords $\kappa$ (\S\ref{sec:action}).
  \item \textbf{Transition} $f_\mathrm{tr}(s_t, a_t) \to s_{t+1}$: tool execution followed by node integration and (optionally) edge inference, producing the updated KG (\S\ref{sec:edge_inference}).
  \item \textbf{Goal test} $\phi(s_t, q) \in \{0, 1\}$: an LLM-judged sufficiency check on whether $\mathcal{G}_t$ contains enough evidence to answer query $q$; termination occurs when $\phi = 1$ or $t = T_{\max}$.
\end{itemize}

Unlike plan-then-execute frameworks, the agent observes $s_t$, selects $a_t$ conditioned on accumulated state, and immediately incorporates environment feedback before selecting $a_{t+1}$. This state-conditioned planning structure is the central mechanism we isolate empirically against PageIndex (\S\ref{sec:main_results}).

\subsection{Environment: Vector-Free Multimodal Document Tree}
\label{sec:env}

The overall architecture is shown in Figure~\ref{fig:pipeline}. The planning loop described in this section (state estimation, action selection, dynamic termination) is architecturally domain-agnostic and applies to any hierarchically structured document corpus. Regulatory documents such as FSARs present the conditions under which this approach is most advantageous over conventional RAG: deep hierarchical structure that chunking destroys, dense cross-references between sections and figures, multimodal evidence (specification tables, engineering drawings) that co-determines answers, and a review process that inherently requires multi-hop evidence gathering with sufficiency judgment. The domain-specific component is the edge ontology (Section~\ref{sec:edge_inference}), which encodes regulatory reasoning relations (\textsc{Satisfies}, \textsc{Violates}); the rest of the architecture transfers directly to other structured document domains.

The environment is represented as a JSON hierarchical tree organized into chapter $\to$ section $\to$ paragraph nodes, preserving the native structure of regulatory documents without any chunking or embedding. To support multimodal reasoning, the system parses the LIST OF FIGURES/TABLES and detects in-text references such as ``Figure 5.1-1,'' attaching figure and table metadata to the corresponding nodes via a \texttt{references} field. This directly addresses the ``figure on different page'' problem, wherein the referencing text and the actual diagram reside on different pages of the PDF.

Rather than relying on dense vector retrieval, the system adopts a vector-free design using BM25Okapi keyword search over the full document tree. Section titles receive a $3\times$ weight boost, and document-length normalization naturally promotes short, focused leaf nodes to higher rankings. At the scale evaluated in this work, the tree spans Ch.01 with 866 nodes (34 figures, 19 tables) and Ch.05 with 26 nodes (29 figures, 30 tables).

\begin{figure*}[t]
  \centering
  \includegraphics[width=\textwidth]{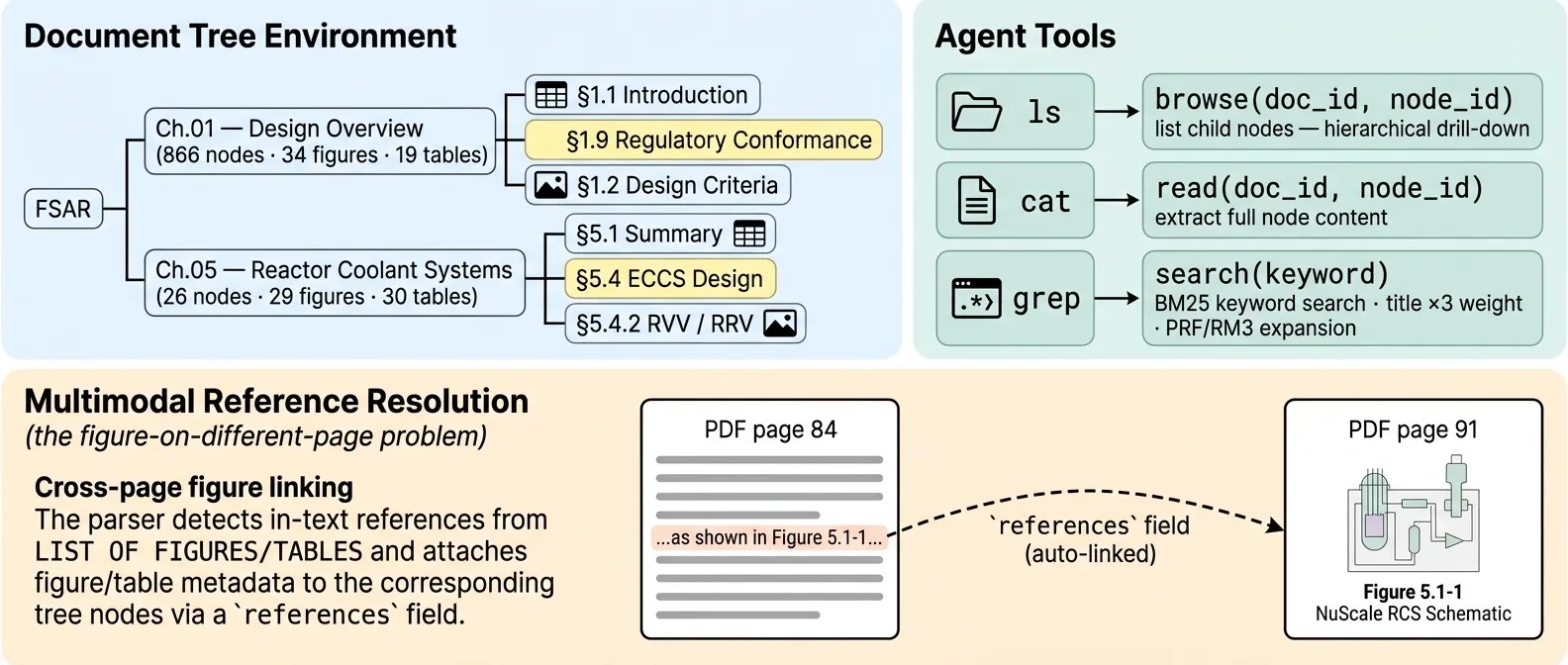}
  \caption{Document tree environment (left), three agent tools (right), and multimodal reference resolution linking in-text references to actual PDF pages (bottom).}
  \label{fig:doc_tree}
\end{figure*}

\subsection{State (Short-Term Memory): Dynamic Sub-KG and Two-Tier Edge Ontology}
\label{sec:state}

The agent state at timestep $t$ is defined as a dynamic knowledge graph $\mathcal{G}_t = (\mathcal{V}_t, \mathcal{E}_t)$. The node set $\mathcal{V}_t$ comprises document sections (evidence nodes) collected through exploration along with their associated multimodal references. The edge set $\mathcal{E}_t$ is governed by a domain-specific two-tier ontology, with edges retained only when confidence is $\geq 0.4$ (empirically set).

The ontology is summarized in Table~\ref{tab:edges}. Tier 1 consists of \emph{structural} edges that organize the exploration trajectory:

\begin{table}[t]
  \caption{Two-tier regulatory edge ontology.}
  \label{tab:edges}
  \begin{center}
          \resizebox{\columnwidth}{!}{
      \begin{tabular}{lp{5cm}}
        \toprule
        \textbf{Edge} & \textbf{Semantics \& Grounding} \\
        \midrule
        \multicolumn{2}{l}{\emph{Tier 1: Structural}} \\
        \textsc{References} & Cross-reference (A cites B) \\
        \textsc{Specifies} & Higher-level elaborated by lower-level~\cite{sysml2014} \\
        \midrule
        \multicolumn{2}{l}{\emph{Tier 2: Semantic}} \\
        \textsc{Satisfies} / \textsc{Violates} & Design meets/fails requirement~\cite{sysml2014,construction_ontology2012} \\
        \textsc{Supports} / \textsc{Contradicts} & Evidence corroborates/refutes claim~\cite{argmining2013,textual_entailment2012} \\
        \textsc{Leads\_To} & Causal chain~\cite{causal_kg2019} \\
        \textsc{Is\_Prerequisite\_Of} & Concept dependency~\cite{prerequisite2017} \\
        \bottomrule
      \end{tabular}}
  \end{center}
\end{table}

\begin{figure}[t]
  \centering
  \includegraphics[width=\columnwidth]{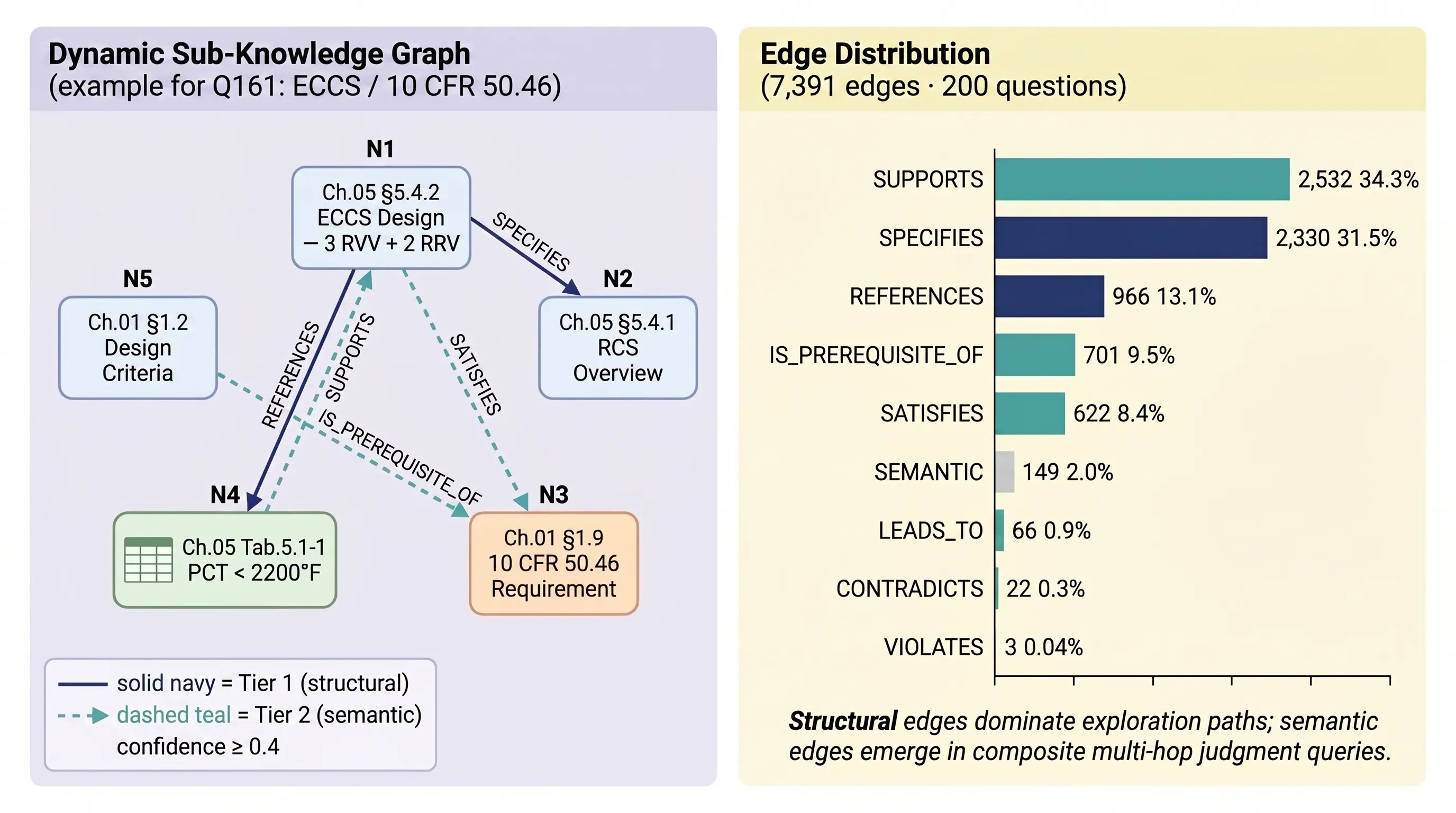}
  \caption{An example Dynamic Sub-Knowledge Graph showing five evidence nodes connected by structural (Tier 1) and semantic (Tier 2) edges. The edge distribution summary (right) shows the prevalence of each edge type across 7{,}391 edges from 200 questions.}
  \label{fig:subkg}
\end{figure}

An example KG with edge distribution is shown in Figure~\ref{fig:subkg}. Empirically, structural edges (\textsc{References}, \textsc{Specifies}) dominate in single-hop factual queries by forming the exploration path, while semantic edges (\textsc{Satisfies}, \textsc{Supports}) emerge in composite multi-hop judgment queries to support regulatory compliance synthesis. In correct answers relative to incorrect ones, \textsc{Supports} appears $+6.8$ percentage points more frequently and \textsc{Satisfies} $+3.2$ percentage points more frequently.

\subsection{Action Planning: LLM-Based Tool Selection}
\label{sec:action}

Rather than precomputing a full retrieval plan, the system performs closed-loop planning. At each hop, the agent observes the current KG state $\mathcal{G}_t$ and decides action $a_t$, with environment feedback (retrieved results) immediately incorporated into the subsequent plan. This constitutes a state-based iterative decision-making structure, distinct from the plan-then-execute separation of APEX-Searcher \cite{apexsearcher2026} and the token-level reactive retrieval of Self-RAG \cite{selfrag2024}. Unlike passive embedding-similarity retrieval in conventional RAG, the LLM actively evaluates the current state and plans the next action.

The agent has access to three tools (Figure~\ref{fig:doc_tree}) that mirror filesystem operations: \textbf{browse} lists the child nodes of a tree node (analogous to \texttt{ls}), \textbf{read} extracts the full content of a specific node (analogous to \texttt{cat}), and \textbf{search} performs BM25-ranked keyword search across all documents (analogous to \texttt{grep}).

A \emph{browse-first} pattern is enforced at Hop 1, where the document structure (table of contents, ToC) is automatically injected so that the agent obtains a global map before searching. This intervention improved single-evidence Context Recall from 0.45 to 0.89. To address vocabulary mismatch, Pseudo-Relevance Feedback (PRF, RM3) automatically expands queries using the top-3 retrieved results at zero additional LLM cost. The agent also maintains a search history to prevent duplicate keyword queries across hops.

\textbf{Dynamic termination} is implemented as a plan sufficiency check beginning at Hop 2: before each hop, the LLM judges whether the current KG already contains sufficient evidence to answer the query. If so, the agent terminates early. This functions as a goal test within the planning loop, automatically calibrating exploration depth to query complexity. Across 200 questions, 33\% of queries terminate early at 1--3 hops (mean 3.4, maximum 4).

\subsection{Post-Retrieval Edge Inference (Optional Component)}
\label{sec:edge_inference}

Edge inference makes explicit the relationships among collected evidence nodes and is performed concurrently with the state transition $f_\mathrm{tr}(\mathcal{G}_t, a_t) \to \mathcal{G}_{t+1}$. The accuracy impact of this component is evaluated in Section~\ref{sec:ablation}; we offer it as an optional module for use cases requiring traceability.

Inference proceeds in two stages. In Stage 1 (Description), the LLM produces a single natural-language sentence describing the relationship between two nodes, without imposing any classification pressure. This follows the free-form relation extraction approach of LightRAG \cite{lightrag2024}; for example: ``The ECCS design of 3 RVV + 2 RRV is configured to meet the acceptance criteria of 10 CFR 50.46.'' In Stage 2 (Ontology Mapping), the free-form description is mapped onto the regulatory domain ontology (\textsc{Satisfies}, \textsc{Violates}, etc.). When no mapping is applicable, the relationship is preserved as \textsc{Semantic}, ensuring no relational information is discarded.

Verification is integrated into the evidence-gathering loop at every hop rather than applied as a post-processing step: immediately after new evidence is retrieved (planning), relationship inference is performed (verification), and the resulting enriched KG state informs the plan sufficiency judgment for the subsequent hop. This design differs from embedding-based implicit relations used in GraphRAG and GWM \cite{gwm2025}; by grounding relationships in explicit LLM-generated natural-language descriptions, the inference results remain human-inspectable.

\subsection{Vision-Augmented Final Answer Generation}
\label{sec:answer}

Multimodal processing is applied exclusively at the final answer generation step, with all intermediate operations (search, plan, infer) remaining text-only. This cost-efficient design avoids the expense of vision API calls during iterative exploration while still enabling visually grounded final answers.

The implementation proceeds in three steps: (1) all Figure/Table references across KG nodes are collected; (2) the corresponding PDF pages are rendered to JPEG using PyMuPDF; and (3) the full text KG context together with the rendered images is passed to the GPT-4.1 vision-language model (VLM) API. For tables specifically, PyMuPDF's \texttt{find\_tables()} function extracts row and column structure directly as structured text, making VLM image processing unnecessary. This approach achieves 86.0\% accuracy on table-only questions, compared to 68.0\% for RAPTOR ($+18$ percentage points).

The complete pipeline is summarized in Algorithm~\ref{alg:pipeline}.

\begin{algorithm}[t]
  \caption{\ours{} Planning Pipeline}
  \label{alg:pipeline}
  \begin{algorithmic}
    \STATE \textbf{Input:} Query $q$, documents $\mathcal{D}$, max hops $T$
    \STATE Initialize $\mathcal{G}_0 \leftarrow \emptyset$; inject ToC at hop 0
    \FOR{$t = 0$ \textbf{to} $T-1$}
      \STATE $a_t \leftarrow \mathrm{LLM}(q, \mathcal{G}_t, \mathrm{ToC})$ \COMMENT{plan actions}
      \STATE Execute $a_t$ via \texttt{browse}/\texttt{read}/\texttt{search}
      \STATE $\mathcal{G}_{t+1} \leftarrow f_\mathrm{tr}(\mathcal{G}_t, a_t)$ \COMMENT{integrate + infer edges (optional)}
      \IF{$t \geq 1$ \textbf{and} SufficientEvidence$(q, \mathcal{G}_{t+1})$}
        \STATE \textbf{break}
      \ENDIF
    \ENDFOR
    \STATE \textbf{return} VisionAugmentedAnswer$(q, \mathcal{G}_\mathrm{final})$
  \end{algorithmic}
\end{algorithm}
            % method.md        (Section 3)
%--- Benchmark (compact for 8-page limit)
\section{Benchmark: Nuclear Regulatory Multi-hop QA}
\label{sec:benchmark}

No existing benchmark combines nuclear regulatory domain, multi-hop reasoning, multimodal evidence (tables and engineering drawings), and regulatory judgment. We survey the most relevant benchmarks along these dimensions in Table~\ref{tab:benchmark_comparison}. FDARxBench \cite{fdarxbench2026} is the most similar (regulatory documents with judgment) but covers only single documents without engineering drawings. MMLongBench-Doc \cite{mmlongbench2024} and M3DocRAG \cite{m3docrag2024} support multimodal understanding but lack regulatory judgment. DesignQA \cite{designqa2024} addresses engineering compliance but not multi-hop reasoning. NuclearQA \cite{nuclearqa2023} targets single-hop factual extraction only.

\begin{table}[t]
  \caption{Existing document QA benchmarks. \benchmark{} is the first to combine all five properties.}
  \label{tab:benchmark_comparison}
  \begin{center}
    \begin{small}
      \resizebox{\columnwidth}{!}{
      \begin{tabular}{lccccc}
        \toprule
        \textbf{Benchmark} & \textbf{M-hop} & \textbf{Tab.} & \textbf{Fig.} & \textbf{X-doc} & \textbf{Judg.} \\
        \midrule
        NuclearQA \cite{nuclearqa2023}     & --  & --  & --      & --  & --  \\
        FDARxBench \cite{fdarxbench2026}   & \checkmark & \checkmark & Part. & --  & \checkmark \\
        MMLongBench \cite{mmlongbench2024} & \checkmark & \checkmark & \checkmark & --  & --  \\
        M3DocRAG \cite{m3docrag2024}       & \checkmark & \checkmark & \checkmark & \checkmark & --  \\
        DesignQA \cite{designqa2024}       & --  & \checkmark & \checkmark & \checkmark & Part. \\
        SEC-QA \cite{secqa2024}            & \checkmark & \checkmark & --  & \checkmark & --  \\
        TAT-QA \cite{tatqa2021}            & \checkmark & \checkmark & --  & --  & --  \\
        \midrule
        \textbf{Ours} & \checkmark & \checkmark & \checkmark & \checkmark & \checkmark \\
        \bottomrule
      \end{tabular}}
    \end{small}
  \end{center}
\end{table}

\subsection{Design and Composition}
\label{sec:benchmark_composition}

We construct 200 questions over NuScale FSAR Chapter 01 (352 pages) and Chapter 05 (160 pages), organized along three orthogonal axes. \textbf{Reasoning type}: factual (70), comparative (65), judgment (65). \textbf{Evidence complexity}: single-evidence (50), multi-evidence (75), cross-document (75). \textbf{Modality}: text-only (80), table-only (50), image-only (30), composite (40). The benchmark contains 357 ground-truth evidence items (152 text, 125 table, 80 figure). This distribution is summarized in Table~\ref{tab:benchmark_matrix}; the judgment $\times$ cross-document cell (35 questions) is the largest, reflecting the core regulatory review task.

\begin{table}[t]
  \caption{Question distribution in \benchmark{} (200 total).}
  \label{tab:benchmark_matrix}
  \begin{center}
    \begin{small}
      \begin{tabular}{lccc|c}
        \toprule
        & \textbf{Single} & \textbf{Multi} & \textbf{Cross-doc} & \textbf{Total} \\
        \midrule
        Factual (70)     & 30 & 25 & 15 & 70 \\
        Comparative (65) & 15 & 25 & 25 & 65 \\
        Judgment (65)    &  5 & 25 & \textbf{35} & 65 \\
        \midrule
        \textbf{Total} & 50 & 75 & 75 & 200 \\
        \bottomrule
      \end{tabular}
    \end{small}
  \end{center}
\end{table}

\subsection{Dual Evaluation Framework}
\label{sec:benchmark_eval}

We adopt dual evaluation: \textbf{RAGAS} \cite{ragas2024} for grounding quality (Faithfulness, Answer Relevancy, Context Recall, Factual Correctness) and \textbf{LLM-as-Judge} \cite{llmasjudge2023} with three independent evaluators, Eval-A (GPT-4-turbo), Eval-B (GPT-4o), and Eval-C (Claude Sonnet 4.5), combined via majority vote. The RAGAS--Judge agreement rate is 66.2\%, with the 34\% disagreement providing complementary information: RAGAS measures grounding fidelity while the Judge captures practical correctness.
         % benchmark.md     (Section 4)
%--- Experiments & Analysis (compact main body; details in Appendix)
\section{Experiments}
\label{sec:experiments}

\subsection{Setup and Baselines}
\label{sec:setup}

All methods use GPT-4.1 for generation with temperature 0 and max\_tokens 300. Our agent is configured with max\_hops$=$4 and top\_k$=$2. We compare against RAPTOR \cite{raptor2024} (recursive summarization tree), HippoRAG \cite{hipporag2024} (hippocampal associative KG), LightRAG \cite{lightrag2024} (dual-level graph + vector DB), GraphRAG \cite{graphrag2024} (community-based local search), and PageIndex \cite{pageindex2024} as an ablation baseline. Full baseline configurations are in Appendix~\ref{app:baselines}.

PageIndex is the critical comparison: it operates over the \textbf{identical} document tree with the same tools, browse-first ToC injection, BM25 configuration, PRF/RM3 query expansion, search history deduplication, and dynamic termination. The only difference is action selection: our system conditions on accumulated KG state; PageIndex selects actions from the query and immediate results alone. The gap therefore isolates state-conditioned planning.

\subsection{Main Results}
\label{sec:main_results}

We report LLM-as-Judge accuracy in Table~\ref{tab:main_results}. Our system (planning only) achieves \textbf{81.5\%} overall. The most important comparison is PageIndex at 43.5\%, isolating a $+38.0$pp planning contribution (McNemar $p<0.001$). Among external baselines, we significantly outperform HippoRAG ($p<0.01$), LightRAG ($p<0.05$), and GraphRAG ($p<0.001$). The $+6.0$pp gap over RAPTOR is consistent across all categories but does not reach significance (McNemar $p=0.11$, $n=200$); however, our system requires zero pre-indexing cost versus RAPTOR's 44 minutes and \$1.4 of offline indexing.

\textbf{Judgment caveat.} 98\% of judgment questions have ``Yes'' as the correct answer (FSARs document compliant designs by definition), so the judgment column has limited discriminative power. Comparative and factual categories provide more meaningful comparisons.

\begin{table*}[t]
  \caption{LLM-as-Judge accuracy on \benchmark{} (200 questions). PageIndex table-only and composite results were not collected in the same evaluation run.}
  \label{tab:main_results}
  \begin{center}
    \begin{small}
      \begin{tabular}{lccccccc}
        \toprule
        \textbf{Method} & \textbf{Overall} & \textbf{Judg.} & \textbf{Comp.} & \textbf{Fact.} & \textbf{Cross-doc} & \textbf{Table} & \textbf{Compos.} \\
        \midrule
        \textbf{Ours (plan.\ only)} & \textbf{81.5} & \textbf{92.3} & \textbf{80.0} & \textbf{72.9} & \textbf{84.0} & \textbf{86.0} & 77.5 \\
        Ours (plan.+edges) & 81.0 & 90.8 & 78.5 & 74.3 & 81.3 & 86.0 & \textbf{85.0} \\
        RAPTOR & 75.5 & 92.3 & 72.3 & 62.9 & 73.3 & 68.0 & 72.5 \\
        LightRAG & 73.0 & 92.3 & 66.1 & 61.4 & 76.0 & 60.0 & 67.5 \\
        HippoRAG & 70.5 & 86.2 & 63.1 & 62.9 & 69.3 & 64.0 & 60.0 \\
        GraphRAG & 49.5 & 61.5 & 49.2 & 38.6 & 37.3 & 42.0 & 47.5 \\
        PageIndex & 43.5 & 61.5 & 44.6 & 25.7 & 45.3 & --- & --- \\
        \bottomrule
      \end{tabular}
    \end{small}
  \end{center}
\end{table*}

\subsection{RAGAS and Efficiency}
\label{sec:ragas}

We report RAGAS metrics in Table~\ref{tab:ragas}. Our system achieves Faithfulness of \textbf{0.93} and Context Recall of \textbf{0.93}, ranking first across all metrics. GraphRAG scores 0.28/0.18, indicating factual loss during community summarization. PageIndex achieves only 0.58 Faithfulness, illustrating the limits of unguided retrieval. The per-reasoning-type breakdown (Table~\ref{tab:ragas_bytype}) confirms this dominance across all three categories: our system leads both Faithfulness and Context Recall on factual, comparative, and judgment questions. GraphRAG's Context Recall of 0.11 on factual questions confirms that specific numerical values are systematically lost during community summarization.

\begin{table}[t]
  \caption{RAGAS comparison across all methods.}
  \label{tab:ragas}
  \begin{center}
    \begin{tabular}{lcccc}
      \toprule
      \textbf{Method} & \textbf{Faith.} & \textbf{AR} & \textbf{CR} & \textbf{FC} \\
      \midrule
      GraphRAG  & 0.28 & 0.59 & 0.18 & 0.32 \\
      PageIndex & 0.58 & 0.77 & 0.66 & 0.30 \\
      RAPTOR    & 0.74 & 0.83 & 0.77 & 0.40 \\
      HippoRAG & 0.76 & 0.83 & 0.76 & 0.37 \\
      LightRAG  & 0.89 & 0.83 & 0.88 & 0.36 \\
      \textbf{Ours} & \textbf{0.93} & \textbf{0.84} & \textbf{0.93} & \textbf{0.42} \\
      \bottomrule
    \end{tabular}
  \end{center}
\end{table}

\begin{table*}[t]
  \caption{RAGAS by reasoning type across all methods. Our system leads Faithfulness and Context Recall on all three reasoning types.}
  \label{tab:ragas_bytype}
  \begin{center}
    \begin{small}
      \begin{tabular}{ll cccccc}
        \toprule
        \textbf{Type} & \textbf{Metric} & \textbf{Ours} & \textbf{LightRAG} & \textbf{RAPTOR} & \textbf{HippoRAG} & \textbf{PageIndex} & \textbf{GraphRAG} \\
        \midrule
        Factual     & Faith. & \textbf{0.92} & 0.91 & 0.81 & 0.84 & 0.58 & 0.26 \\
        Factual     & CR     & \textbf{0.92} & 0.86 & 0.75 & 0.79 & 0.57 & 0.11 \\
        Comparative & Faith. & \textbf{0.92} & 0.88 & 0.68 & 0.70 & 0.54 & 0.32 \\
        Comparative & CR     & \textbf{0.91} & 0.89 & 0.74 & 0.71 & 0.67 & 0.18 \\
        Judgment    & Faith. & \textbf{0.97} & 0.89 & 0.74 & 0.74 & 0.63 & 0.27 \\
        Judgment    & CR     & \textbf{0.96} & 0.91 & 0.82 & 0.76 & 0.76 & 0.25 \\
        \bottomrule
      \end{tabular}
    \end{small}
  \end{center}
\end{table*}

\textbf{Efficiency.} We report indexing and per-query costs across methods in Table~\ref{tab:efficiency}. Our system requires 8--20 min and \$4.1 for pre-indexing, but higher per-query cost (\$0.215, 93s) due to multi-hop exploration. Total cost for 200 questions is $\sim$\$46 vs.\ $\sim$\$2.3 for RAPTOR, yielding a per-accuracy-point cost of \$0.56/\%p. Dynamic termination partially mitigates this: 33\% of queries terminate early, with costs ranging from \$0.03 (1-hop factual) to \$0.29 (4-hop cross-document judgment). For regulatory applications, where human reviewers spend hours per question, the \$46 total for 200 questions is negligible compared to manual review cost.

\begin{table*}[t]
  \caption{Efficiency comparison across methods. Per-query cost extrapolated from 5-question sample with $\pm$10--15\% variance.}
  \label{tab:efficiency}
  \begin{center}
    \begin{small}
      \begin{tabular}{l cc cc cc cc}
        \toprule
        & \multicolumn{2}{c}{\textbf{Indexing}} & \multicolumn{2}{c}{\textbf{Per-query}} & \multicolumn{2}{c}{\textbf{Total (200Q)}} & \multicolumn{2}{c}{\textbf{Summary}} \\
        \cmidrule(lr){2-3} \cmidrule(lr){4-5} \cmidrule(lr){6-7} \cmidrule(lr){8-9}
        \textbf{Method} & Time & Cost & Time & Tokens & Time & Cost & Acc. & \$/\%p \\
        \midrule
        \textbf{Ours} & 8--20\,min & \$4.1 & 93\,s & 86K & $\sim$320\,min & $\sim$\$46 & \textbf{81.5\%} & \$0.56 \\
        RAPTOR   & $\sim$44\,min & $\sim$\$1.4 & 1.8\,s & 2K   & $\sim$50\,min & $\sim$\$2.3 & 75.5\% & \$0.03 \\
        LightRAG & 52\,min       & $\sim$\$7.0 & $\sim$5\,s & 24K  & $\sim$69\,min & $\sim$\$17  & 73.0\% & \$0.23 \\
        HippoRAG & 29\,min       & $\sim$\$4.3 & $\sim$3\,s & 2.6K & $\sim$39\,min & $\sim$\$5.4 & 70.5\% & \$0.08 \\
        GraphRAG & 40\,min       & $\sim$\$3.3 & 5.2\,s & 693   & $\sim$57\,min & $\sim$\$3.7 & 49.5\% & \$0.07 \\
        \bottomrule
      \end{tabular}
    \end{small}
  \end{center}
\end{table*}

%-----------------------------------------------------------------------
\section{Analysis}
\label{sec:analysis}

\subsection{Component Ablation (10Q)}
\label{sec:ablation_10q}

To isolate each component's contribution, we remove one at a time from the full system on a 10-question subset spanning diverse reasoning types (Table~\ref{tab:ablation10q}). Only the full system achieves 10/10, and each removal fails on a distinct question type: Vision RAG removal drops Q101 (table/comparative) and Q131 (composite/comparative), with Faithfulness falling from 0.96 to 0.83; Edge inference removal drops Q058 (seismic scope boundary) due to missing \textsc{Violates} edge but saves 66\% of cost; Browse-first removal drops Q191 (image/judgment/cross) due to lost structural orientation. At $n=10$, differences are indicative rather than statistically conclusive, motivating the 200Q scale-up for the edge inference question.

\begin{table}[t]
  \caption{10-question component ablation. Each removal fails on a distinct question type.}
  \label{tab:ablation10q}
  \begin{center}
    \resizebox{\columnwidth}{!}{
      \begin{tabular}{lcccccc}
        \toprule
        \textbf{Variant} & \textbf{Judge} & \textbf{Faith.} & \textbf{AR} & \textbf{CR} & \textbf{Time} & \textbf{Cost} \\
        \midrule
        \textbf{full} & \textbf{10/10} & \textbf{0.96} & 0.84 & 0.95 & 104s & \$0.216 \\
        no\_vision & 8/10 & 0.83 & 0.82 & 0.92 & 98s & \$0.196 \\
        no\_edges & 9/10 & 0.93 & 0.84 & \textbf{1.00} & \textbf{34s} & \textbf{\$0.073} \\
        no\_browse\_first & 9/10 & 0.97 & 0.79 & 0.94 & 91s & \$0.180 \\
        \bottomrule
      \end{tabular}}
  \end{center}
\end{table}

\subsection{Ablation: Edge Inference at Scale (200Q)}
\label{sec:ablation}

Since edge inference accounted for 65\% of per-query cost in the 10Q ablation, we scale this comparison to all 200 questions (Table~\ref{tab:ablation200q}).

\begin{table}[t]
  \caption{Edge inference ablation (200 questions).}
  \label{tab:ablation200q}
  \begin{center}
    \begin{tabular}{lcc}
      \toprule
      \textbf{Metric} & \textbf{Plan.+Edges} & \textbf{Plan.\ Only} \\
      \midrule
      3-Judge Accuracy & 81.0\% & \textbf{81.5\%} \\
      Faithfulness & \textbf{0.930} & 0.897 \\
      Context Recall & \textbf{0.930} & 0.919 \\
      Cost / question & \$0.215 & \textbf{\$0.076} \\
      Time / question & 115.3s & \textbf{47.5s} \\
      \bottomrule
    \end{tabular}
  \end{center}
\end{table}

\textbf{Key finding: Planning is the primary driver.} Removing edge inference does not reduce accuracy; it provides traceability (human-readable paths like ``Section A \textsc{Satisfies} Regulation B'') but no accuracy gain. Planning mechanisms alone, namely browse-first (CR 0.45$\to$0.89), dynamic termination (33\% early), and state-conditioned selection (PageIndex $+38.0$pp), suffice to outperform all baselines. This parallels APEX-Searcher \cite{apexsearcher2026} and PRISM \cite{prism2025} in scope but diverges in method: both require RL/SFT training, while our system achieves equivalent planning in a \textbf{training-free} setting.

\subsection{Edge Distribution and Case Studies}
\label{sec:edge_dist}

Across 200 questions, 7{,}391 edges are generated: \textsc{Supports} (34.3\%), \textsc{Specifies} (31.5\%), \textsc{References} (13.1\%), \textsc{Is\_Prerequisite\_Of} (9.5\%), \textsc{Satisfies} (8.4\%), with \textsc{Violates} at only 0.04\% (3 instances). Semantic edges correlate with correctness: \textsc{Supports} $+6.8$pp and \textsc{Satisfies} $+3.2$pp more frequent in correct answers.

\textbf{Why \textsc{Violates} appears in a certified document.} The FSAR has already been certified by the U.S. Nuclear Regulatory Commission (NRC), so the emergence of \textsc{Violates} edges warrants explanation. All three instances capture \textbf{scope boundary exclusions} rather than design deficiencies. This reveals a regulatory reasoning capability absent from standard RAG:

\begin{itemize}
  \item \textbf{Q058 (Seismic scope, $\times$2):} Both \textsc{Violates} edges (confidence 0.85, 0.90) identify that non-safety-related systems (Chilled Water, Condensate Storage) are intentionally placed outside Seismic Category I classification. The FSAR explicitly justifies this: ``failure of non-safety systems, structures, and components (SSCs) does not affect safety-related SSCs.'' The agent marks these scope boundaries in the KG, so that a human reviewer can immediately see which requirements do not apply. Without this marking, such information would require manual cross-referencing across chapters. The distinction matters for regulatory review: ``requirement does not apply'' and ``requirement is not satisfied'' carry different licensing consequences, and a representation that does not separate them structurally shifts the burden of distinguishing onto the human reviewer.

  \item \textbf{Q176 (Partial conformance, $\times$1):} The \textsc{Violates} edge (confidence 0.85) captures that NuScale's integrated Steam Generator (SG) design eliminates the traditional containment bypass problem but introduces a leakage detection limitation: the system cannot distinguish identified from unidentified leakage, resulting in partial conformance with Design-Specific Review Standard (DSRS) 15.6.5. This is the space between full \textsc{Satisfies} and full \textsc{Violates}, the kind of state where regulatory review typically requires additional engineering analysis or compensatory measures. The agent represents the state as a typed edge with confidence 0.85; the baselines we evaluate in \S\ref{sec:main_results} produce no equivalent typed annotation.
\end{itemize}

These cases show that while edge inference does not improve accuracy (Table~\ref{tab:ablation200q}), it provides concrete regulatory value: (1) explicit scope boundary identification, marking where requirements apply and do not apply; (2) partial conformance that mirrors actual regulatory judgment; and (3) auditable reasoning paths satisfying the traceability requirements of 10 CFR 50 Appendix B. The rarity of \textsc{Violates} (3 of 7{,}391 edges, 0.04\%) itself serves as a quality signal consistent with a certified document. For practical use, these three typed edges identify the locations a human reviewer can focus on without re-reading every natural-language answer.

The 66.2\% RAGAS--Judge agreement leaves 67 disagreement cases that decompose into two qualitatively distinct categories. \textbf{29 cases (RAGAS Good + Judge X)}: answers grounded in the KG (Faith $\geq 0.8$, CR $\geq 0.8$) but penalized for wording differences from the reference. Q019 is representative: reference ``the pressurizer volume is 578 ft$^3$,'' agent ``the pressurizer region volume is 578 ft$^3$ and the cylindrical section is 487 ft$^3$,'' RAGAS Faith 1.00 / CR 1.00, Judge X. These reflect benchmark-level evaluation strictness, not regulatory error. \textbf{38 cases (RAGAS Bad + Judge O)}: answers Judge marks correct despite low Context Recall, meaning the agent answered from knowledge not fully captured by the retrieved KG. This is the deployment-relevant category: $38/200 = 19\%$ of correct answers cannot be fully verified by inspecting the KG alone, and a reviewer needing audit-grade reasoning should flag this fraction for separate verification. The two categories motivate dual evaluation: RAGAS measures grounding, Judge measures correctness, and either alone misclassifies cases that matter for safety review.

\subsection{Failure Mode Taxonomy}
\label{sec:failure_modes}

We classify the 38 incorrect answers (Judge X) by RAGAS scores, using the same Faith $\geq 0.8$ and CR $\geq 0.8$ thresholds as \S\ref{sec:ragas}.

\textbf{Expression / reasoning} (29/38, 76.3\%): well-grounded by both metrics, Judge X. The dominant pattern is wording mismatch with the reference, as in Q019 above. The penalty is for added detail, not regulatory error.

\textbf{Hallucination} (2/38, 5.3\%): Faith $<$ 0.5. The answer asserts content the KG does not support.

\textbf{Retrieval miss} (1/38, 2.6\%): CR $<$ 0.5. Retrieval missed evidence the reference relies on.

\textbf{Mixed} (6/38, 15.8\%): intermediate Faith and CR.

Excluding the wording-mismatch category, the deployment-relevant failure rate is $9/200 = 4.5\%$. Hallucination at scale is $2/200 = 1.0\%$. These are the rates a reviewer should expect to verify against external evidence.

\subsection{Self-Assessment Calibration}
\label{sec:calibration}

Dynamic termination (\S\ref{sec:action}) lets the agent decide when its evidence is sufficient. Whether the early-stopping signal is reliable determines whether early termination introduces a deployment risk. We stratify accuracy by hop count: early-terminated queries (1--3 hops) are the high-confidence subset; full-budget queries (4 hops) are the low-confidence subset.

\begin{table}[t]
  \caption{Accuracy by hop count.}
  \label{tab:hop_accuracy}
  \begin{center}
    \begin{small}
      \begin{tabular}{lccc}
        \toprule
        \textbf{Bucket} & \textbf{Hops} & $n$ & \textbf{Accuracy} \\
        \midrule
        Early & 1 & 11 & 100.0\% \\
        Early & 2 & 13 & 84.6\% \\
        Early & 3 & 45 & 82.2\% \\
        Full  & 4 & 129 & 78.3\% \\
        \midrule
        \textbf{Early total} & 1--3 & 69 & \textbf{85.5\%} \\
        \textbf{Full total}  & 4    & 129 & 78.3\% \\
        \bottomrule
      \end{tabular}
    \end{small}
  \end{center}
\end{table}

Early-terminated queries reach 85.5\% accuracy; full-budget queries reach 78.3\%. Higher accuracy on the high-confidence subset is the calibration signature: the early-stopping signal is reliable. Question difficulty confounds the comparison (simple factual queries terminate at hop 1, cross-document judgment queries use 4 hops), but the ordering rules out the failure mode in which the agent quits too early on hard cases. That mode would invert the ordering.

       % experiment.md    (Sections 5 & 6)
%--- Limitations (promoted to its own section, before Conclusion)
\FloatBarrier
\section{Limitations}
\label{sec:limitations}

\paragraph{System.}
Our system underperforms RAPTOR on text-only questions (76.2\% vs.\ 80.0\%, $-3.8$pp); adding summary nodes to the tree is a potential improvement. Per-query cost (\$0.215, 93\,s) is 42$\times$ RAPTOR's cost but justified for regulatory use where a human reviewer requires hours per question; the system's 94.3\% accuracy on the 35 judgment$\times$cross-document questions, the core regulatory review task, justifies the cost premium where it matters most. The RAPTOR gap ($+6.0$pp) cannot be asserted as a reliable improvement at the current sample size (McNemar $p=0.11$). A follow-reference tool for directly navigating ``see Table 5.1-1''-style pointers remains unimplemented. The max\_hops$=$4 ceiling is used by 67\% of queries; whether increasing the budget would improve accuracy on the most complex questions remains an open question.

\paragraph{Benchmark.}
The benchmark exhibits five structural limitations identified during evaluation: (1) \emph{Factual Correctness ceiling} ($\sim$0.42 across all methods) reflects single-perspective expected answers rather than retrieval failure; (2) \emph{judgment polarity bias}: 98\% ``Yes'' answers, as FSARs document by definition compliant designs; (3) \emph{limited evidence depth}: 66\% of questions are effectively 2-hop; (4) \emph{document coverage imbalance}: Ch.01 uses only 19\% of its pages; (5) \emph{no external validation}: the benchmark is self-designed, with mitigating factors including the three-axis orthogonal design, 3-evaluator majority voting, and uniform comparison across 5 methods under identical conditions.

\paragraph{Edge ontology coverage.}
Our eight-relation ontology is chosen for FSAR conformance review and does not cover relations relevant to other regulatory frameworks, such as cross-jurisdictional applicability, temporal validity, conditional applicability, and engineering exception precedents. The three \textsc{Violates} instances we report come from a single FSAR corpus; whether similar typed-edge patterns appear in other regulatory domains is unknown. Deploying the system to a different framework requires domain-expert ontology curation and human oversight.

%--- Conclusion
\section{Conclusion}
\label{sec:conclusion}

This paper frames multi-hop regulatory document review as a planning problem and instantiates it as an LLM-based agent that operates over a vectorless document tree with a dynamic knowledge graph as state. On a 200-question NuScale FSAR benchmark, the system reaches 81.5\% accuracy with RAGAS Faithfulness 0.93, outperforming GraphRAG, HippoRAG, and LightRAG, and matching RAPTOR while eliminating offline indexing cost. The $+38.0$pp gap over PageIndex isolates state-conditioned planning as the primary accuracy driver. A 200-question ablation finds edge inference contributes no accuracy gain at 2.8$\times$ cost, but produces typed regulatory relations: 3 \textsc{Violates} edges among 7{,}391 (0.04\%) make scope-bounded inapplicability and partial conformance explicit, the audit form that 10 CFR 50 Appendix B requires. Failure mode analysis shows 29 of 38 errors are wording mismatch rather than regulatory error, leaving a deployment-relevant failure rate of 4.5\%; hop-stratified accuracy (85.5\% early vs.\ 78.3\% full) confirms that the early-stopping mechanism is well-calibrated. The architecture (document-as-environment, action interface, KG state, dynamic termination) transfers to any domain whose review process requires multi-hop evidence gathering and explicit conformance judgment, with the edge ontology as the domain-specific component.
        % conclusion.md    (Section 7)

\section*{Impact Statement}
The primary goal of this work is to improve the efficiency and accuracy of
agent-based machine learning systems. We anticipate no immediate negative
societal impacts arising uniquely from our contributions beyond those generally
associated with the deployment of large language model based agents. On the
contrary, by reducing the computational cost of multi-hop reasoning over large
document collections, our approach can help make sophisticated agentic workflows
more accessible and environmentally efficient in settings where compute is a
limiting factor. In safety-critical domains such as nuclear regulatory review,
we emphasize that the system is intended to assist, not replace, expert human
judgment, and that its outputs should remain subject to domain-expert oversight.

\section*{Acknowledgements}
This work was supported by the Substantiation Support Program, through the
Korea Innovation Foundation funded by the Ministry of Science and ICT
(No.~76170-26).

% === Bibliography ===
\bibliography{paper}
\bibliographystyle{icml2026}

% === Appendix (does not count toward 8-page limit) ===
%--- Appendix (unlimited pages per LM4Plan CFP)
\clearpage
\onecolumn
\appendix
\section*{Appendix}

\section{Baseline Configurations}
\label{app:baselines}

The configuration used for each baseline is detailed in Table~\ref{tab:baselines_full}. All methods share the same generation LLM (GPT-4.1, temperature 0, max\_tokens 300) and operate on the same 200-question benchmark over NuScale FSAR Chapters 01 and 05. The configurations below reflect each method's published defaults, tuned only where the original settings were incompatible with our document corpus (e.g., chunk size adjusted to accommodate FSAR section lengths).

\textbf{RAPTOR} constructs a recursive summarization tree using 100-token leaves, producing 3{,}422 leaves from our corpus, with retrieval via the \texttt{collapse\_tree} strategy under a 2{,}000-token budget. \textbf{HippoRAG} extracts a hippocampal associative knowledge graph from 1{,}000-character passages (1{,}080 in total) and retrieves via Personalized PageRank combined with dense similarity (top-10). \textbf{LightRAG} performs three-pass entity-relation extraction over 1{,}200-token chunks (266 chunks), using hybrid retrieval of top-40 entities plus top-20 chunks. \textbf{GraphRAG} builds a community graph over 1{,}200-token chunks (267 chunks) and retrieves via local search at community level 2. \textbf{PageIndex} shares our system's tree environment and tools (browse, read, search) exactly, differing only in action-selection logic (no KG state accumulation), and thus serves as the single-variable ablation for state-conditioned planning.

\begin{table}[h]
  \caption{Baseline configurations. All methods use GPT-4.1 for generation.}
  \label{tab:baselines_full}
  \begin{center}
    \resizebox{\textwidth}{!}{
    \begin{tabular}{llllp{5cm}}
      \toprule
      \textbf{Method} & \textbf{Paper} & \textbf{Core Mechanism} & \textbf{Chunking} & \textbf{Retrieval Config} \\
      \midrule
      RAPTOR & \cite{raptor2024} & Recursive summ.\ tree & 100 tok, 3{,}422 leaves & collapse\_tree, 2K token budget \\
      HippoRAG & \cite{hipporag2024} & PPR + hippocampal KG & 1{,}000 chars, 1{,}080 pass. & PPR + dense, top-10 \\
      LightRAG & \cite{lightrag2024} & Dual-level graph + vector DB & 1{,}200 tok, 266 chunks & hybrid, top-40 ent.\ + 20 chunks \\
      GraphRAG & \cite{graphrag2024} & Community-based local search & 1{,}200 tok, 267 chunks & local search, community\_level=2 \\
      PageIndex & \cite{pageindex2024} & Vectorless tree (no planning) & None (tree nodes) & BM25 + browse/read/search, max 4 hops \\
      \bottomrule
    \end{tabular}}
  \end{center}
\end{table}

\section{RAGAS: Our System by Reasoning Type}
\label{app:ragas_detail}

The cross-method per-type breakdown is shown in the main body (Table~\ref{tab:ragas_bytype}). Here we provide our system's decomposition across all four RAGAS metrics (Faithfulness, Answer Relevancy, Context Recall, Factual Correctness) by reasoning type, giving a fuller picture of where the planning loop is strongest and where it struggles.

\begin{table}[h]
  \caption{RAGAS metrics for our system by reasoning type.}
  \label{tab:ragas_ours_detail}
  \begin{center}
    \begin{tabular}{lcccc}
      \toprule
      \textbf{Metric} & \textbf{Overall} & \textbf{Fact.} & \textbf{Comp.} & \textbf{Judg.} \\
      \midrule
      Faithfulness & \textbf{0.93} & 0.92 & 0.92 & \textbf{0.97} \\
      Answer Rel.  & \textbf{0.84} & 0.85 & 0.78 & \textbf{0.89} \\
      Context Rec. & \textbf{0.93} & 0.92 & 0.91 & \textbf{0.96} \\
      Factual Corr.& 0.42 & 0.35 & \textbf{0.49} & 0.41 \\
      \bottomrule
    \end{tabular}
  \end{center}
\end{table}

Two patterns emerge. First, \textbf{judgment questions dominate across three of four metrics} (Faith.\ 0.97, AR 0.89, CR 0.96), indicating that the planning loop is particularly effective at assembling complete evidence chains for regulatory conformance reasoning, the scenario most aligned with multi-hop state-conditioned exploration. Second, \textbf{Factual Correctness is the only metric where judgment does not lead}: the 0.42 overall FC reflects the wording-sensitive nature of this metric (exact string match against the expected answer) rather than a retrieval failure. A closer look at flagged cases shows that FC penalizes correct answers phrased differently than the reference (e.g., ``vertical helical once-through SG with 1{,}380 tubes'' vs.\ the reference's ``helical coil SG integrated within RPV''). Both responses are factually correct; the metric penalizes the paraphrase. This is a benchmark-level limitation (single-perspective reference answers) rather than a model failure.

\section{Per-Question Cost Breakdown}
\label{app:efficiency}

To illustrate the dynamic termination mechanism concretely, we profiled five representative questions drawn from different points in the three-axis taxonomy. Token counts were measured with \texttt{tiktoken} (\texttt{o200k\_base} encoding) over the full LLM inputs (retrieved context, system prompts, and agent reasoning) and outputs, and costs were computed using GPT-4.1 pricing (\$2/M input tokens, \$8/M output tokens). Because the LLM is non-deterministic even at temperature 0 (due to backend batching), node and edge counts can vary slightly between runs; the numbers below are from a single dedicated profiling run.

\begin{table}[h!]
  \caption{Per-question cost breakdown (5-question profiling sample). Illustrates dynamic termination: simple queries (Q001) cost \$0.03, complex queries (Q191) cost \$0.29.}
  \label{tab:perquestion}
  \begin{center}
    \begin{tabular}{llcccrc}
      \toprule
      \textbf{QID} & \textbf{Type} & \textbf{Hops} & \textbf{Nodes} & \textbf{Edges} & \textbf{Tokens} & \textbf{Cost} \\
      \midrule
      Q001 & factual / single     & 1 & 4  & 0  & 9{,}395   & \$0.03 \\
      Q071 & comparative / single & 4 & 19 & 63 & 124{,}264 & \$0.30 \\
      Q131 & composite / cross    & 4 & 11 & 34 & 75{,}213  & \$0.18 \\
      Q161 & judgment / multi     & 4 & 17 & 65 & 104{,}463 & \$0.25 \\
      Q191 & image / cross        & 4 & 19 & 81 & 117{,}023 & \$0.29 \\
      \bottomrule
    \end{tabular}
  \end{center}
\end{table}

The breakdown demonstrates that cost scales approximately with question complexity: the single-hop factual query Q001 consumes less than 10K tokens and costs \$0.03, while cross-document queries (Q071, Q131, Q161, Q191) use the full 4-hop budget and cost between \$0.18 and \$0.30. Node counts also correlate with complexity, ranging from 4 nodes (single factual) to 19 nodes (multi-hop cross-document). Edge counts scale super-linearly with node counts, reflecting the pairwise nature of the two-stage edge inference. Total 200-question costs reported in the main body (Table~\ref{tab:efficiency}) are extrapolated from this sample with an estimated $\pm$10--15\% variance; a larger profiling run would tighten these estimates but would not change the qualitative conclusion that per-query cost is dominated by a small number of 4-hop complex queries.

\section{10Q Ablation: Per-Question Judge Detail}
\label{app:10q_ablation}

The 10-question ablation summary table is shown in the main text (Table~\ref{tab:ablation10q}). Here we provide the per-question O/X detail across the four variants (\texttt{full}, \texttt{no\_vision}, \texttt{no\_edges}, \texttt{no\_browse\_first}). The 10 questions were selected to span the three-axis taxonomy, including at least one question from each reasoning type and each modality.

\begin{table}[h!]
  \caption{Per-question 3-Judge detail across ablation variants.}
  \label{tab:judge_detail}
  \begin{center}
    \begin{tabular}{llcccc}
      \toprule
      \textbf{QID} & \textbf{Type} & \textbf{full} & \textbf{no\_vis} & \textbf{no\_edg} & \textbf{no\_brw} \\
      \midrule
      Q001 & fact/single       & O & O & O & O \\
      Q010 & fact/multi        & O & O & O & O \\
      Q031 & fact/single/table & O & O & O & O \\
      Q058 & fact/cross        & O & O & \textbf{X} & O \\
      Q071 & comp/single       & O & O & O & O \\
      Q101 & comp/cross/table  & O & \textbf{X} & O & O \\
      Q131 & comp/cross/comp   & O & \textbf{X} & O & O \\
      Q161 & judg/multi        & O & O & O & O \\
      Q176 & judg/cross        & O & O & O & O \\
      Q191 & judg/cross/image  & O & O & O & \textbf{X} \\
      \bottomrule
    \end{tabular}
  \end{center}
\end{table}

\textbf{Robustness vs.\ diagnostic questions.} Six of the ten questions (Q001, Q010, Q031, Q071, Q161, Q176) are answered correctly by all four variants, indicating that the baseline system, even with any single component removed, handles single-hop factual queries, multi-hop factual queries, and judgment queries over well-structured evidence paths. These six \emph{robustness} questions confirm that the planning loop as a whole is resilient to individual component ablation for straightforward cases. The remaining four questions are \emph{diagnostic}: each is failed by exactly one variant, revealing which component is the critical dependency for that question type.

\textbf{Failure modes by component.} \textbf{Vision RAG removal} drops Q101 (table/comparative) and Q131 (composite/comparative). These are the only two questions requiring numerical content from figure-rendered tables; without vision processing the system cannot recover the relevant cells. Faithfulness also drops from 0.96 to 0.83, reflecting ungrounded answers when tabular evidence is withheld. \textbf{Edge inference removal} drops Q058 (seismic scope boundary), where the \textsc{Violates} edge was required to mark non-safety-related systems as outside Category I scope. This failure does not replicate at the 200Q scale: both variants answer Q058 correctly in the main 200Q evaluation, indicating the 10Q Q058 result reflects execution-level variance rather than a systematic dependency. \textbf{Browse-first removal} drops Q191 (image/judgment/cross), where the agent, lacking the table-of-contents injection at Hop 1, selects poorly-targeted sections and never recovers the cross-document evidence chain.

\textbf{Why scale up only edge inference to 200Q.} Each component is responsible for a distinct failure mode at the 10Q level, supporting the complementarity claim. However, Q058 was the only failure that we suspected might be a sample-size artifact (because the \textsc{Violates} edge is a rare edge type, it is plausible that the dependency does not generalize). Scaling edge inference to 200Q confirms this suspicion: the effect is reliably null (Table~\ref{tab:ablation200q}), while the vision and browse-first effects continue to manifest (Q101, Q131, Q191 all remain diagnostic at larger scales). This selective scale-up also reflects practical constraints: the 200Q full ablation consumes roughly 200$\times$ the compute of the 10Q pilot, so we prioritized the ablation with the strongest a priori cost-accuracy hypothesis (edge inference accounts for 65\% of per-query cost).

\end{document}